\documentclass[aps,pre,reprint,superscriptaddress,longbibliography]{revtex4-2}

\usepackage{amsmath,amssymb,bm}
\usepackage{graphicx}
\usepackage{booktabs}
\usepackage{microtype}
\usepackage{needspace}
\usepackage{placeins}
\usepackage{xcolor}

\newcommand{\one}{\bm{1}}
\newcommand{\R}{\mathbb{R}}
\newcommand{\dd}{\mathrm{d}}
\newcommand{\cL}{\mathcal{L}}
\newcommand{\cQ}{\mathcal{Q}}
\newcommand{\diag}{\operatorname{diag}}

\newtheorem{proposition}{Proposition}

\begin{document}

\title{Reciprocity Separates Gradient Flow from Rotation in Conservative Physical Learning}

\author{Ruiwu Niu}
\email{rniu@hksyu.edu}
\affiliation{Department of Data Science and Digital Innovation, Hong Kong Shue Yan University}

\author{Xiaowen Bi}
\email{xiaowenbi@bnbu.edu.cn}
\affiliation{Faculty of Science and Technology, Beijing Normal-Hong Kong Baptist University}

\author{Micha\"el Antonie van Wyk}
\affiliation{School of Electrical and Information Engineering, University of the Witwatersrand, Johannesburg, South Africa}

\date{August 30, 2026}

\begin{abstract}
Physical learning lets a trainable material or network use its own physical
response to
carry error signals, reducing the need for a separately programmed backward
computation.  We ask what determines whether such a system follows conventional
gradient descent or evolves along a genuinely different learning trajectory.  Our
canonical model is
a directed layered transport network in which every node redistributes a fixed
amount of flow, so learning preserves positivity and total mass.  In this model,
conservation constrains only the allowable learning directions.  Within the
matched response class studied here, adjoint matching gives the physical output
response a symmetric form.  Non-negative mode-wise feedback then produces a
reciprocal closed-loop response and a reweighted gradient flow.  Adding an
antisymmetric boundary component makes the closed-loop
response rotational: the learning path can turn while the error driving that
update still decreases at that moment.  Turning is not automatically
beneficial.  Its finite-step
effect is set by local curvature, and its accumulated effect also depends on step
selection and on the new states visited along the path.  Numerical consistency
checks reproduce the exact response structure, predict the sign of the local effect across new
network families, and show how trajectory drift can negate a local advantage.
These results separate the roles of conservation, reciprocity, and
nonreciprocity in physical learning.
\end{abstract}

\maketitle

\section{Introduction}

In physical learning, the response of the trainable medium itself performs part of
the training computation.  Equilibrium propagation and coupled learning use this
idea to turn global task information into local parameter
updates~\cite{scellier2017equilibrium,stern2021supervised,dillavou2022laboratory}.
Related schemes use chemical feedback, analytical circuit solvers, or local
evolution rules to avoid an explicitly assembled digital backward
pass~\cite{anisetti2023learning,ezraty2026harnessing,lin2026resistive}.  Experiments
and simulations have also resolved physical traces of learned functions and the
loss of previously learned tasks during sequential
training~\cite{stern2024physical,stern2025physical,guzman2025microscopic,
ibrahim2026sequential}.  These developments raise a structural question within
the matched response class considered here: when can a physical update be
understood as gradient descent in a state-dependent geometry, and must
reciprocity be broken for learning to acquire a genuinely new direction of
motion?

Consider a trainable network that routes a normalized flow from input to output.
Each node may redistribute the incoming flow but may neither create nor destroy it,
and learning modifies these routing fractions in response to an output error.
Conservation therefore determines where the parameters are allowed to move.  It
does not determine how they move within that constrained space.  The same
mass-preserving state space can support downhill, rotational, or mixed dynamics.
Replicator dynamics, natural gradient, mirror descent, and
exponentiated-gradient updates provide a geometric language for describing this
constrained motion~\cite{shahshahani1979framework,amari1998natural,raskutti2015information},
but locality and normalization alone do not determine which dynamics arises.

A useful distinction within the matched response class considered here is between
the physical response and the closed-loop learning response.  The former maps an
applied boundary potential to an output change, while the boundary law maps the
current error to that potential.  Their composition determines whether the
error-to-output dynamics are reciprocal.  With adjoint matching and non-negative
spectral feedback, this composition can change the metric and conditioning of
gradient descent but cannot provide an independent circulation.  This restricted
connection is consistent with generalized gradient structures in nonequilibrium
thermodynamics~\cite{mielke2016onsager} and with recent circuit analyses of
gradient-based physical rules and higher-order coupled-learning
corrections~\cite{mcginnis2026coercivity,mcginnis2026conservation}.

An antisymmetric component in the closed-loop response introduces a transverse
direction.  Directed and
active interactions are known to generate oscillatory modes and new collective
dynamics~\cite{fruchart2021nonreciprocal,suchanek2023entropy,
beuria2026topological}, while non-gradient rotational components can either help or
hinder learning depending on their structure~\cite{ninou2025curl}.  Here, the
antisymmetric component is placed in the closed-loop learning response, not merely
in the material being trained.  Contrastive learning can train metamaterials with
nonreciprocal target responses~\cite{du2026metamaterials}; our question is instead
what happens when the error-to-output learning dynamics acquire an antisymmetric
component.

Our argument proceeds in four steps.  First, conservation defines the admissible
parameter manifold but does not make the learning field a gradient.  Second,
adjoint matching gives the physical response a Gram form at any network depth,
while non-negative spectral feedback yields a reciprocal closed-loop response and
preconditioned gradient flow.  Third, a minimal three-output boundary controller
provides a rotational closed-loop response while preserving output mass and
instantaneous dissipation of the selected-sample error.  Fourth, a finite-step
curvature coefficient characterizes the local effect, while registered
finite-horizon accounting decomposes the observed final loss gap.  Together, these
steps provide two response results, a finite-step expansion, and an accounting
identity for comparing registered finite trajectories.

\begin{figure*}[t]
  \includegraphics[width=\textwidth]{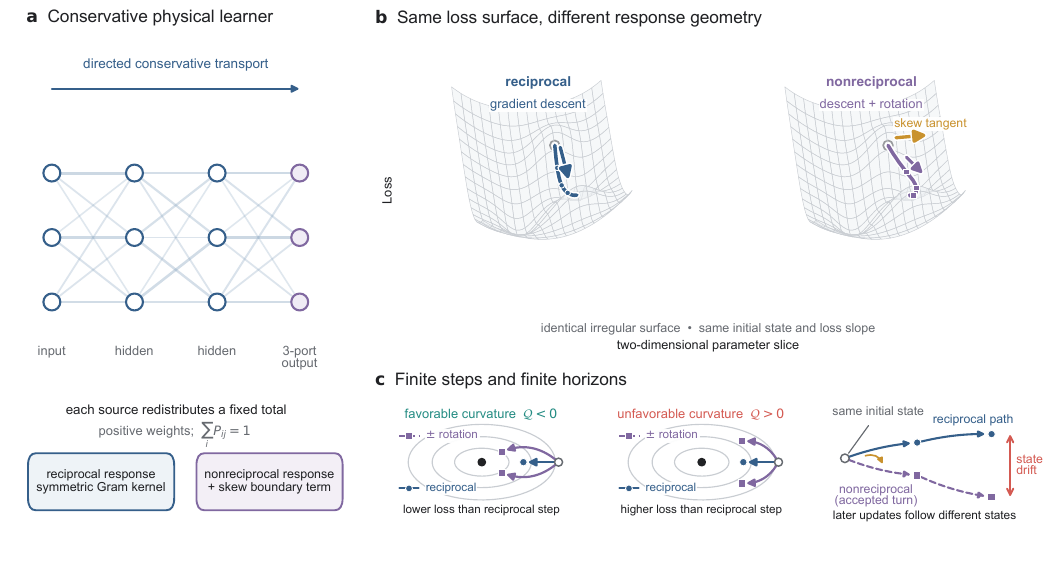}
  \caption{\textbf{Conceptual geometry of conservative physical learning.}
  (a) A positive directed layered network redistributes a fixed total flow.
  Reciprocity refers to the closed-loop learning response, not to the direction of
  the forward transport edges.  Adjoint matching produces a symmetric Gram
  physical response; non-negative spectral feedback keeps the closed loop
  symmetric, while a skew boundary term adds an antisymmetric component.  (b) On
  a schematic two-dimensional parameter slice, an irregular selected-sample loss
  landscape is shown in two identical copies for legibility.  The
  reciprocal trajectory follows the local gradient, whereas the nonreciprocal
  trajectory descends while turning around landscape features.  At their common
  initial state, the skew component is tangent to the selected-sample level set
  and therefore leaves its first-order loss slope unchanged; the foreground blue
  and gold
  arrows mark the local descent and skew directions, respectively.  The turning
  freedom is not itself a performance advantage.
  (c) For a finite normalized update, averaging the two rotation orientations can
  either lower or raise the post-step loss, depending on local curvature.  The
  right schematic compares a reciprocal trajectory, shown as a blue solid line
  with circles, with a nonreciprocal trajectory after an accepted turn, shown as
  a purple dashed line with squares; both begin from the same state.  Across
  repeated updates, the resulting separation of visited states produces a drift
  term that can offset a local gain.  The diagram summarizes the theory and
  contains no experimental data.}
  \label{fig:concept}
\end{figure*}

The numerical tests follow the same sequence.  They verify the matched-response
factorization and rotational modes, test a fixed curvature-based sign criterion on
disjoint
network families, and account for the final loss gap along paired reciprocal and
nonreciprocal trajectories.  Figure~\ref{fig:concept} is a conceptual roadmap
rather than a data figure.  Read from left to right, panel (a) separates
conservation from reciprocity at the network level, panel (b) shows that
nonreciprocity introduces a transverse direction to local descent, and panel (c)
shows why the value of that
direction depends first on finite-step curvature and then on the states visited
later.

\section{Conservative layered networks}
\label{sec:model}

\subsection{Product-simplex parameters}

We use a directed layered network as a canonical conservative learner.  Each
column of a layer matrix contains the fractions of flow sent from one source to
the nodes in the next layer.  Non-negativity prevents negative flow, while a unit
column sum prevents its creation or loss.  Training therefore changes
redistribution fractions rather than the total amount of flow.  The trainable
state consists of the positive probability vectors that form the
column-stochastic layer matrices,
\begin{equation}
  \theta=(p^1,\ldots,p^m),\qquad
  p^a\in\Delta_{d_a}^{\circ},
  \label{eq:state}
\end{equation}
where $\Delta_d^{\circ}=\{p\in\R_{>0}^d:\one^{\mathsf T}p=1\}$ is the
interior of a probability simplex.  The full parameter space is a product of
these simplices: one for every source column.  For any column $p$, define
\begin{equation}
  Q(p)=\diag(p)-pp^{\mathsf T}.
  \label{eq:q}
\end{equation}
$Q(p)$ is positive semidefinite, annihilates $\one$, and maps an arbitrary
local score, interpreted as a preference for changing the routing fractions,
to a velocity tangent to the simplex.  Let $s^a\in\R^{d_a}$ denote the local
score vector for block $a$; its components rank the preference for increasing
the corresponding routing fractions.  A block-local law can therefore be
written as
\begin{equation}
  \dot p^a=Q(p^a)s^a,
  \qquad \one^{\mathsf T}\dot p^a=0.
  \label{eq:replicator}
\end{equation}
For one block, write the components as $p_i$ and $s_i$.  Given a step size
$\eta>0$, the normalized exponential step
\begin{equation}
  p_i^+(\eta)=
  \frac{p_i\exp(\eta s_i)}{\sum_jp_j\exp(\eta s_j)}
  \label{eq:retraction}
\end{equation}
preserves positivity and column mass.  It is an entropic mirror step for a
prescribed score.  Thus $Q(p)$ and the normalized step determine where learning
is allowed to move; the score field determines whether that motion is downhill,
rotational, or a combination of the two.

Let $F(\theta)$ stack the outputs for all samples, and let $Y$ stack the
corresponding targets.  Define
\begin{equation}
  e=Y-F(\theta),\qquad
  \cL(\theta)=\frac{1}{2}\lVert e\rVert^2,
  \qquad J=D_\theta F.
  \label{eq:loss}
\end{equation}
Here, $D_\theta F$ denotes the derivative of $F$ with respect to the full
parameter vector $\theta$, so $J$ is the Jacobian that maps parameter
perturbations to output changes.  For every sample $s$, the inputs and targets
are non-negative and have unit mass:
$\one^{\mathsf T}x_s=\one^{\mathsf T}y_s=1$.  Column-stochastic propagation
preserves this normalization, so every sample residual belongs to the zero-sum
output subspace.  Consequently, three outputs leave two independent error
coordinates and can support the rotation shown in Fig.~\ref{fig:concept}b.

We collect the local mobility operators in
\begin{equation}
  M(\theta)=\operatorname{blkdiag}_{a}
  \bigl(\rho_a Q(p^a)\bigr),\qquad \rho_a>0.
  \label{eq:M}
\end{equation}
The coefficient $\rho_a$ is the positive mobility assigned to block $a$.  It
sets the block's relative response rate without changing its conservation
constraint.
For a boundary potential $v$, the canonical matched update transports the
boundary signal back through the adjoint Jacobian:
\begin{equation}
  \dot\theta=MJ^{\mathsf T}v,
  \qquad \dot F=J\dot\theta=:Kv.
  \label{eq:response}
\end{equation}
Here, $K$ is the physical output response: it maps an applied boundary potential
$v$ to the instantaneous change in all predictions.  If a boundary controller
sets $v=\mathcal B e$, the error induces the closed-loop learning response
$\mathcal R_{\mathcal B}:=K\mathcal B$, so that
$\dot F=\mathcal R_{\mathcal B}e$ and
$\dot e=-\mathcal R_{\mathcal B}e$.  We distinguish symmetry of the physical
response $K$ from symmetry of the closed-loop response $\mathcal R_{\mathcal B}$.
Below, reciprocal learning refers to symmetry of the latter on the identifiable
output subspace.  The model assumption in Eq.~\eqref{eq:response} is matched
adjoint transport; its Gram consequence is derived in
Sec.~\ref{sec:reciprocal}.  Neither symmetry requires the forward transport edges
to be bidirectional.  A device realization would additionally have to implement
the matched forward and adjoint transport.

\section{Conservation and integrability}
\label{sec:integrability}

Conservation provides a geometric baseline.  It confines each velocity to the
tangent space of a simplex, but it does not by itself impose a scalar potential.
A gradient structure additionally requires the local score differences to satisfy
an integrability condition.

The following geometric representation makes this distinction explicit.
Consider a smooth vector field $V$ satisfying
$\one^{\mathsf T}V^a=0$ on each simplex block.  In the interior, any such tangent
field can be written blockwise as
\begin{equation}
  V^a=Q(p^a)s^a,
  \qquad s_i^a=V_i^a/p_i^a,
  \label{eq:universal}
\end{equation}
where $s^a$ is defined up to an additive multiple of $\one$.  Thus every smooth
conservative tangent field has a block-replicator representation.  This
representation enforces the constraint, but it does not by itself imply a loss
function or a gradient structure.

To test whether the score field is integrable on the simplex, remove one redundant
coordinate from each block and form the reduced score one-form
\begin{equation}
  \omega_s=
  \sum_{a}\sum_{i<d_a}
  \bigl(s_i^a-s_{d_a}^a\bigr)\,\dd p_i^a.
  \label{eq:oneform}
\end{equation}
Here $\dd p_i^a$ denotes an infinitesimal change in the independent simplex
coordinate $p_i^a$.  The eliminated coordinate is fixed by conservation, so
$\dd p_{d_a}^a=-\sum_{i<d_a}\dd p_i^a$.  The score difference
$s_i^a-s_{d_a}^a$ compares the local drive on coordinate $i$ with that on the
reference coordinate $d_a$.  Because
$Q(p^a)\one=0$, adding the same constant to every component of
$s^a$ leaves the physical velocity $V^a$ unchanged.  Only score differences
are therefore identifiable, and Eq.~\eqref{eq:oneform} takes $s_{d_a}^a$ as the
reference score.
On a simply connected interior domain, $V$ is a Shahshahani natural-gradient
field if and only if the associated one-form is closed:
\begin{equation}
  \dd\omega_s=0.
  \label{eq:closed}
\end{equation}
Equivalently, the reduced score differences are integrable and derive from a
single scalar potential.  Thus integrability imposes a condition beyond
conservation.  For example, the rock--paper--scissors replicator field obeys
Eq.~\eqref{eq:replicator} exactly but has periodic interior orbits, so it cannot
be a gradient flow of a single-valued potential.  Related potential--harmonic
decompositions appear in finite games~\cite{candogan2011flows}.  Conservation
therefore fixes the feasible motion while leaving circulation possible.  This
observation separates the two geometric conditions used below.  The next section
adds response assumptions that exclude such circulation within the matched class.

\Needspace{12\baselineskip}
\section{Adjoint-matched response and spectral feedback}
\label{sec:reciprocal}

We now restrict attention to the adjoint-matched response model.  In this class,
the physical response $K$ has a symmetric positive-semidefinite Gram form.
Whether the closed-loop response $\mathcal R_{\mathcal B}$ is gradient-like also
depends on the boundary controller.

Consider a depth-$L$ linear network with positive column-stochastic layer matrices,
\begin{equation}
  x_s^{(\ell+1)}=P^{(\ell)}x_s^{(\ell)},
  \qquad \one^{\mathsf T}P^{(\ell)}=\one^{\mathsf T}.
  \label{eq:deep}
\end{equation}
Let $p_i^{(\ell)}$ be column $i$ of layer $\ell$, and let
$R^{(\ell)}=P^{(L-1)}\cdots P^{(\ell+1)}$ denote the downstream map from layer
$\ell+1$ to the output.  The
Jacobian block for sample $s$, layer $\ell$, and column $i$ is proportional to
$x_{s,i}^{(\ell)}R^{(\ell)}$.

Under the adjoint-matched update in Eq.~\eqref{eq:response}, the physical response
is a symmetric positive-semidefinite Gram operator at every depth.  Direct block
multiplication gives the layer-resolved response
\begin{align}
  K_L&=J_LM_LJ_L^{\mathsf T}=K_L^{\mathsf T}\succeq0,
  \label{eq:factorization}\\
  K_{rs}^{(L)}&=
  \sum_{\ell=0}^{L-1}\rho_\ell\sum_i
  x_{r,i}^{(\ell)}x_{s,i}^{(\ell)}
  R^{(\ell)}Q\!\left(p_i^{(\ell)}\right)R^{(\ell)\mathsf T}.
  \label{eq:blocks}
\end{align}
Equation~\eqref{eq:blocks} resolves the Gram operator into contributions from
individual layers and source columns, including cross-sample coupling through the off-diagonal blocks
$K_{rs}^{(L)}$.  Depth changes the spectrum, rank, and conditioning of the
response while preserving this Gram structure.

Suppose now that the boundary potential is generated spectrally from the physical
response,
\begin{equation}
  v=r(K)e,
  \qquad r(\lambda)\geq0
  \quad \text{for }\lambda\in\sigma(K).
  \label{eq:spectral}
\end{equation}
Here $r(\cdot)$ is a non-negative scalar function applied to the eigenvalues of
$K$, and $\lambda$ denotes one such eigenvalue.  The notation $\sigma(\cdot)$ is
the spectrum operator, so $\sigma(K)$ is the set of all eigenvalues of $K$.
Equivalently, if
$K=\sum_{\lambda\in\sigma(K)}\lambda P_\lambda$, where $P_\lambda$ is the
orthogonal projector onto the corresponding eigenspace, then
$r(K)=\sum_{\lambda\in\sigma(K)}r(\lambda)P_\lambda$.  Thus $r$ sets the
feedback strength along each response mode.
This family contains direct feedback, Moore--Penrose repair, and
leakage-regularized feedback.

\begin{proposition}[Spectral feedback]
For the boundary law in Eq.~\eqref{eq:spectral}, the closed-loop response
$\mathcal R_{\mathrm R}=Kr(K)$ is symmetric and positive semidefinite.  The loss
is non-increasing, and on the identifiable parameter subspace the induced
parameter dynamics are a positive-semidefinite preconditioned gradient flow.
\end{proposition}

Because $K$ and $r(K)$ share an orthogonal eigenbasis,
\begin{equation}
  \dot\cL=-e^{\mathsf T}Kr(K)e
  =-\!\sum_{\lambda\in\sigma(K)}
  \lambda r(\lambda)\lVert P_\lambda e\rVert^2\leq0.
  \label{eq:dissipation}
\end{equation}
Define $\psi(0)=0$, $\psi(\lambda)=r(\lambda)/\lambda$ for $\lambda>0$, and
\begin{equation}
  H=MJ^{\mathsf T}\psi(K)JM.
  \label{eq:H}
\end{equation}
Then $H=H^{\mathsf T}\succeq0$ and, on the identifiable parameter subspace
consisting of the directions that change the outputs, we have
\begin{equation}
  \dot\theta=-H\nabla_\theta\cL.
  \label{eq:preconditioned}
\end{equation}
The symmetry of $\mathcal R_{\mathrm R}$ excludes an antisymmetric circulation,
while its positive semidefiniteness makes error-aligned feedback dissipative.
Changing the spectral function $r$ can alter the metric and conditioning of
descent, but it cannot create an independent rotational mode within this feedback
family.  A regularized inverse
$v=(K+\mu I)^{-1}e$, for example, gives a metric damped Gauss--Newton direction.
The proposition classifies the geometry of the learning rule; implementation
costs such as time, communication, precision, and energy are separate design
questions.

\section{Minimal nonreciprocal escape}
\label{sec:escape}

To leave the gradient class, the closed-loop response needs room to turn.  A single
two-output residual has only one independent mass-conserving direction, so it can
move only forward or backward along a line.  Three outputs produce a
two-dimensional zero-sum plane, the smallest single-sample space in which a
genuine rotation can occur.  With a batch, sample mixing can supply the additional
dimension even when each sample has two outputs.

For the three-output construction define
\begin{equation}
  P=I-\frac13\one\one^{\mathsf T},\qquad
  C^{\mathsf T}=-C,\qquad C^2=-P,
  \label{eq:PC}
\end{equation}
and the boundary mixer
\begin{equation}
  T_{\alpha,\Omega}=\alpha P+\Omega C,
  \qquad \alpha>0.
  \label{eq:mixer}
\end{equation}
$P$ projects onto the zero-sum error plane, while $C$ acts as a quarter-turn on
that plane.  The coefficient $\alpha$ controls inward relaxation and $\Omega$
controls rotation.

For a selected three-output sample $\star$, write
$J_\star=D_\theta F_\star$, $e_\star=y_\star-F_\star$,
$K_\star=J_\star M J_\star^{\mathsf T}$, and
$\cL_\star=\lVert e_\star\rVert^2/2$.  Assume that
$\operatorname{range}(K_\star)=\one^\perp$.  For this construction,
Eq.~\eqref{eq:response} is applied to the selected map:
$\dot\theta=MJ_\star^{\mathsf T}v$ and $\dot F_\star=K_\star v$.  Choose the
boundary controller
\begin{equation}
  \begin{aligned}
  \mathcal B_{\alpha,\Omega}&=K_\star^\dagger T_{\alpha,\Omega},\\
  v&=\mathcal B_{\alpha,\Omega}e_\star,
  \end{aligned}
  \label{eq:nonreciprocal-v}
\end{equation}
which induces the closed-loop response
\begin{equation}
  \begin{aligned}
  \mathcal R_{\alpha,\Omega}
  &=K_\star\mathcal B_{\alpha,\Omega}=T_{\alpha,\Omega},\\
  \dot e_\star&=-\mathcal R_{\alpha,\Omega}e_\star,
  \qquad \dot\cL_\star=-\alpha\lVert e_\star\rVert^2.
  \end{aligned}
  \label{eq:spiral}
\end{equation}

\begin{proposition}[Three-port rotational witness]
The closed-loop response in Eq.~\eqref{eq:spiral} preserves output mass and
decreases the instantaneous selected-sample squared error.  When $\Omega\neq0$,
the residual linearization has the identifiable eigenvalues
$-\alpha\pm i\Omega$ and therefore cannot coincide, in a neighborhood containing
the stationary point, on the identifiable two-dimensional quotient, with a smooth
positive-definite Riemannian gradient flow.
Two outputs cannot support such a single-sample linear rotation.
\end{proposition}

The physical response $K_\star$ remains symmetric, while the boundary controller
makes $\mathcal R_{\alpha,\Omega}$ nonreciprocal through its skew part $\Omega C$.
The symmetric component pulls the error inward and the skew component turns it
sideways, producing the descending spiral in Fig.~\ref{fig:concept}b.  Its
complex pair supplies the local gradient-flow obstruction.  This explicit
closed-loop construction uses effective active boundary feedback; a topology-local
realization remains open, as discussed in Sec.~\ref{sec:discussion}.

\section{Finite steps and finite horizons}
\label{sec:finite}

Leaving the gradient class does not by itself improve learning.  A continuous
trajectory may reduce loss at every instant, while a finite normalized update
also samples the curvature of the constrained parameter-to-output map.  The
relevant question is whether a rotational displacement lands above or below the
corresponding reciprocal step.

At the same pre-step state, the three-port controller produces the block scores
\begin{align*}
 s^{0,a}&=\rho_a\bigl[J_\star^{\mathsf T}
 K_\star^\dagger\alpha Pe_\star\bigr]^a, &
 s^{1,a}&=\rho_a\bigl[J_\star^{\mathsf T}
 K_\star^\dagger Ce_\star\bigr]^a.
\end{align*}
Thus $s^0$ is the reciprocal score and $s^1$ is the unit-coefficient rotational
score.  Their tangent vectors are
\begin{equation}
 U=\bigl(Q(p^a)s^{0,a}\bigr)_a,
 \qquad V=\bigl(Q(p^a)s^{1,a}\bigr)_a.
 \label{eq:UV}
\end{equation}
Apply Eq.~\eqref{eq:retraction} with score
$s^0+\sigma\Omega s^1$, where $\sigma\in\{-1,+1\}$.  Define $G=JV$ and
\begin{equation}
  H_V=JB+D^2F(\theta)[V,V],
  \label{eq:HV}
\end{equation}
where $B$ is the second derivative of the exponential retraction along $V$.
Here $J$ and $e$ remain the full-dataset Jacobian and residual, so the expansion
measures the dataset-loss effect of a score generated by the selected sample.

\begin{proposition}[Finite-step expansion]
At an interior state, fix the reciprocal and rotational scores during the step
and assume that $F$ is three times continuously differentiable.  Averaging the
two rotation orientations cancels terms that are odd in the orientation.  The
second-order orientation-even term relative to the reciprocal step is
$\eta^2\Omega^2\cQ$, where $\cQ$ can be positive or negative.
\end{proposition}

For the mean loss $\bar{\cL}=\cL/N$, we have
\begin{equation}
  \frac{\bar{\cL}_+(\eta)+\bar{\cL}_-(\eta)}{2}
  -\bar{\cL}_0(\eta)
  =\eta^2\Omega^2\cQ+O(\eta^3),
  \label{eq:secondorder}
\end{equation}
with
\begin{equation}
  \cQ=\frac{1}{2N}
  \left(\lVert G\rVert^2-e^{\mathsf T}H_V\right).
  \label{eq:Q}
\end{equation}
The displacement term is non-negative, whereas the residual-aligned path
curvature can dominate it with either sign.  A negative $\cQ$ means that the mean
of the clockwise and counterclockwise steps has lower post-step loss than the
reciprocal update; a positive $\cQ$ means that it has higher loss.  Two distinct
strictly interior one-layer configurations give $\cQ=-0.9271875$ and
$\cQ=0.026666\ldots$, respectively.
These opposite signs confirm that the finite-step effect can be locally favorable
or unfavorable.  Thus nonreciprocity provides an additional degree of freedom
whose finite-step value is determined by local geometry, as depicted in
Fig.~\ref{fig:concept}c.

A favorable local sign does not guarantee a cumulative advantage.  Once a
rotational update changes the state, subsequent updates are evaluated at different
locations.  We therefore use a registered coupling between a nonreciprocal
trajectory $\theta_t^{\mathrm{NR}}$ and a native reciprocal trajectory
$\theta_t^{\mathrm R}$.  The pair shares its initial state, data order, sample
schedule, orientation schedule, and declared step-control rule.

\paragraph{Registered finite-horizon accounting.}
For this coupling and the reference increments defined in
Appendix~\ref{app:ledger}, adding and subtracting the orientation average, the
step-matched reciprocal increment, and the native reciprocal increment gives the
exact identity
\begin{equation}
  \cL(\theta_T^{\mathrm{NR}})-\cL(\theta_T^{\mathrm R})
  =\sum_{t=0}^{T-1}
  \left(\mathsf E_t+\mathsf H_t+\mathsf S_t+\mathsf D_t\right).
  \label{eq:ledger}
\end{equation}
Here, $\mathsf E_t$ is the mean effect of the two rotation orientations,
$\mathsf H_t$ is the additional effect of the realized clockwise or
counterclockwise choice, $\mathsf S_t$ records the difference in step policy, and
$\mathsf D_t$ records the difference between native reciprocal progress evaluated
at the two current states.  These quantities are defined relative to the registered
reference policy.  The equality is an exact accounting identity, not a unique
causal decomposition.  Fair random handedness can cancel $\mathsf H_t$
conditionally when orientation is chosen after the gate and step are fixed.  The
remaining terms record the local orientation-even effect, the step-policy
difference, and the separation of the visited states under this coupling.

\section{Numerical tests and reproducibility}
\label{sec:numerics}

The numerical tests follow the four-part sequence rather than treating final
accuracy as a single verdict.  We first test the structural identities and their
conditioning, then the local curvature sign rule on new task families, and
finally the registered finite-horizon accounting along coupled trajectories.
Throughout this section, network depth $L$ refers to the number of trainable
propagation matrices $P^{(0)},\ldots,P^{(L-1)}$ between input and output.
All plotted networks have three outputs and hidden layers of width three, so the
depth comparisons change the number of successive redistribution stages rather
than the width of each stage.

\textit{Numerical protocol.}  All checks use deterministic seeds and the
normalized exponential update, which preserves positive column mass.
Layer-product derivatives are compared with complex-step or centered finite
differences.  Reciprocal and nonreciprocal trajectories share their initial state,
data, sample order, orientation schedule, and declared step-control rule.  Trials,
rather than individual updates, are the independent experimental units.
Independent verification programs check the stored primitives and identities.
Proofs, algorithms, full-precision values, and the experimental ledger are
provided in the Supplemental Material.

\subsection{Structural identities, conditioning, and local scaling}

The joint response $K$ acts on the predictions of all samples considered
together.  Its numerical rank counts the response eigenmodes that remain above
the registered $10^{-10}$ tolerance, and therefore measures how many independent
joint output directions can be distinguished at finite precision.  In the
registered three-source, three-output linear family, the full connected rank is
six because each of the three source columns has two independent
mass-conserving output directions.  The condition number of the positive
spectrum is the ratio between the strongest and weakest retained response modes.
A large condition number indicates that some directions respond far more weakly
than others, making inversion and numerical identification fragile.

The structural experiment uses 25 task families at depths $2$, $3$, $4$, and
$6$, giving 100 connected joint response operators, 400 rotation-response checks,
and 900 eight-step trajectory comparisons.  The maximum factorization residual is
$1.21\times10^{-16}$, the maximum trajectory discrepancy is
$2.22\times10^{-16}$, and the maximum complex-mode residual is
$3.93\times10^{-15}$.  It also reproduces instantaneous dissipation and both
signs of Eq.~\eqref{eq:Q}; the fitted log--log slopes are $2.000$ and $1.999$ for
the negative- and positive-$\cQ$ examples.

\begin{figure*}[t]
  \includegraphics[width=\textwidth]{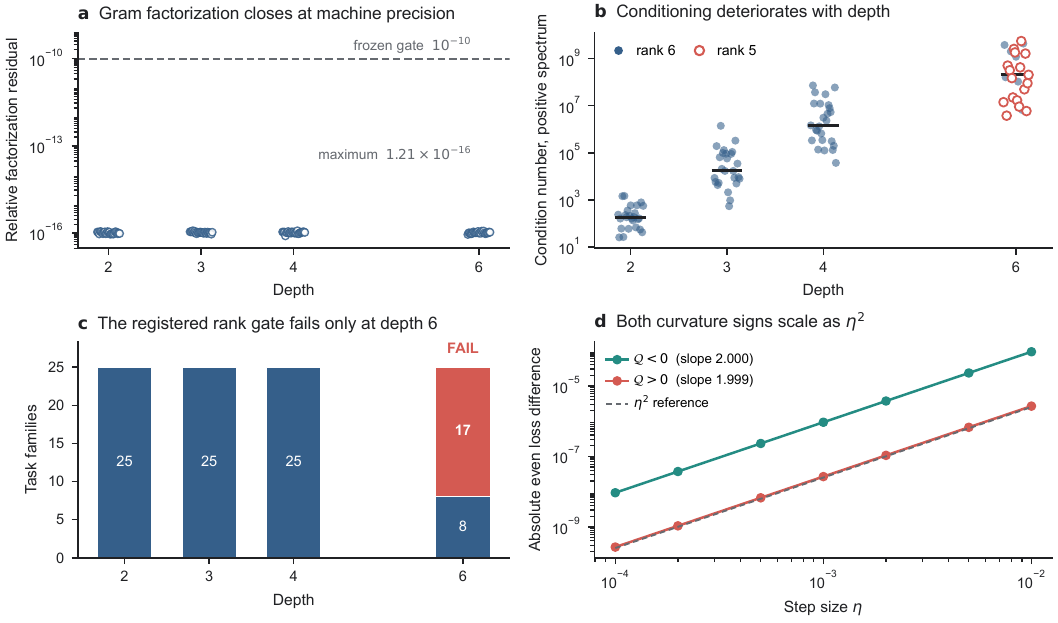}
  \caption{\textbf{Structural tests separate exact response geometry from
  finite-precision identifiability.}  Depth is the number of trainable propagation
  matrices.  Numerical rank counts response modes above the registered tolerance;
  rank six is full rank for the connected three-source, three-output family.
  (a) Relative difference between the explicit layer formula and
  $K=JMJ^{\mathsf T}$ for 25 task families at each depth; the dashed line is the
  preregistered $10^{-10}$ tolerance.  (b) Ratio of the strongest to weakest
  retained response mode.  Horizontal marks show depth-wise medians, and
  filled/open symbols distinguish numerical rank six/five.  (c) Counts of rank-six
  and rank-five operators.  Seventeen of 25 depth-$6$ operators lose one
  numerically identifiable direction, so the universal rank gate remains FAIL
  even though the factorization in (a) still closes.  (d) Orientation-averaged
  loss difference for explicit examples with
  $\cQ<0$ and $\cQ>0$.  Both follow the predicted quadratic dependence on step
  size; reference lines show slope two.  The rank failure concerns finite-precision
  identifiability, whereas panel (d) tests the local curvature expansion.}
  \label{fig:structural}
\end{figure*}

Figure~\ref{fig:structural} asks two separate questions: whether the matched
physical-response factorization is algebraically correct, and whether all of its response modes
remain numerically usable.  Panel (a) compares the explicit layer formula with
$K=JMJ^{\mathsf T}$; residuals near $10^{-16}$ show agreement at machine
precision for every depth.  Panel (b) shows that the same operators become
progressively ill-conditioned, with the weakest positive mode approaching the
finite-precision floor.  Panel (c) counts the consequence.  Every operator at
depths $2$, $3$, and $4$ retains all six modes, but 17 of 25 depth-$6$ operators
retain only five.  In those cases, one expected joint output mode is too weak
to distinguish under the registered tolerance.  The FAIL label therefore rejects
the preregistered claim of full numerical rank for every tested operator.  It
does not reject the Gram factorization, symmetry, or positive semidefiniteness,
all of which still pass in the same depth-$6$ cases.  Practically, it warns that
deep matched physical responses may require regularization because inverse feedback can
amplify roundoff or noise along weak modes.  Panel (d) addresses a separate local
prediction and confirms that examples with either sign of $\cQ$ follow the
predicted $\eta^2$ finite-step scaling.

\subsection{Predicting the local sign in unseen network families}

\begin{figure*}[t]
  \includegraphics[width=\textwidth]{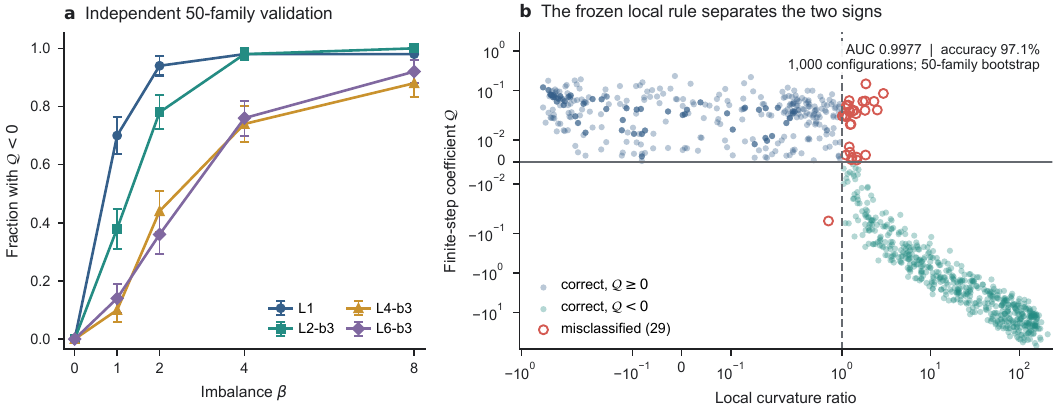}
  \caption{\textbf{Disjoint validation of the local curvature sign rule.}
  Here $\cQ<0$ means a locally lower orientation-averaged one-step loss than the
  reciprocal update, not a guaranteed cumulative advantage.
  (a) Fraction of 50 new task families with $\cQ<0$ for one-layer and
  width-three networks of depths $2$, $4$, and $6$.  The routing concentration
  $\beta$ ranges from uniform columns at zero to strongly concentrated columns at
  eight; error bars show binomial standard errors over task families.  (b) Exact
  full-dataset coefficient versus the selected-sample curvature ratio
  $\chi_s$.  The frozen decision threshold is one.  Filled colors show the true
  sign of $\cQ$, and open red circles mark the 29 misclassified configurations.
  AUC and accuracy intervals resample the 50 task families rather than the 1000
  correlated configuration rows.}
  \label{fig:threshold}
\end{figure*}

Equation~\eqref{eq:Q} compares the squared first-order output displacement with
residual-aligned curvature.  The frozen predictor measures this competition on
the selected sample $s$ through
\begin{equation}
  \chi_s=\frac{e_s^{\mathsf T}H_{V,s}}
  {\lVert G_s\rVert^2},
  \qquad \chi_s>1\ \Longrightarrow\ \text{predict }\cQ<0.
  \label{eq:local-ratio}
\end{equation}
The threshold of one means that curvature is predicted to dominate displacement.
For this test, $\cQ<0$ means that the mean of the two opposite rotational steps
has lower one-step loss than the reciprocal step from the same state.  It is a
local comparison and does not by itself predict the final training loss.
The validation uses 50 new task families, four architectures, and five routing
imbalance values.  In Fig.~\ref{fig:threshold}, the labels L1 and L$d$-b3 denote
one layer and a depth-$d$ network with width-three hidden layers, respectively.
The imbalance parameter
$\beta$ scales column logits before normalization: $\beta=0$ gives uniform
routing, while larger values concentrate flow onto fewer routes.

Figure~\ref{fig:threshold} asks whether one selected-sample measurement predicts
the sign of the full-dataset coefficient in 1000 previously unused
configurations.  Panel (a) plots the fraction of task families with $\cQ<0$.
It shows that locally favorable rotational steps become more common as routing
becomes more imbalanced across all four architectures.  The favorable fraction
rises from zero at $\beta=0$ to $94.5\%$ at $\beta=8$ (95\% task-family interval
$[91.0\%,97.5\%]$).  Panel (b) shows a sharp separation around the fixed threshold
of one: points with $\chi_s>1$ are predicted to lie below $\cQ=0$, while points
with $\chi_s\leq1$ are predicted to lie on or above it.  The predictor reaches an
AUC of $0.998$ with interval
$[0.996,0.999]$ and an accuracy of $97.1\%$ with interval
$[96.2\%,98.0\%]$.  The 29 marked errors show that the local diagnostic is highly
informative but is not identical to the full-dataset coefficient.  Analytic and
finite-step signs agree in all configurations.

\subsection{Why a favorable local turn may disappear over time}

The finite-horizon experiment compares three policies.  The native reciprocal
control uses the full admissible step for the reciprocal score.  The step-matched
nonreciprocal policy reserves step capacity for a possible rotational component,
including updates for which the curvature gate is inactive.  The adaptive policy
restores the native reciprocal step envelope when rotation is switched off and
reserves capacity only when a turn is accepted.  The plotted step scale
$s\in\{0.1,0.2\}$ is the numerator of the normalized envelope
$\eta=s/\max\{1,\max_i u_i-\min_i u_i\}$, where $u$ is the applied score vector.

The cumulative step-policy term records the loss difference associated with the
two admissible steps when both updates are evaluated at the same state.  The
state-drift term records the difference in subsequent reciprocal progress after
the policies have reached different states.  Under the registered decomposition,
these terms separate step allocation from differences associated with the visited
states.

\begin{figure*}[t]
  \includegraphics[width=\textwidth]{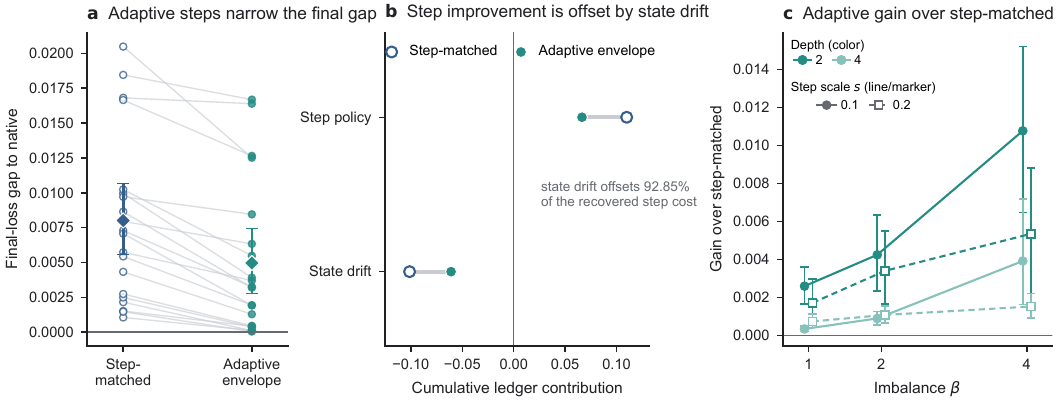}
  \caption{\textbf{Registered finite-horizon accounting separates local gain and
  state drift.}  (a) Trial-level final-loss gaps relative to the native reciprocal
  learner.  Thin lines connect the step-matched and adaptive nonreciprocal
  policies within each trial after averaging the 12 conditions and two handedness
  replicates; diamonds and bars show the mean and 95\% cluster-bootstrap interval
  across 20 trials.  Positive values mean higher final loss than the native
  reciprocal control.  (b) The two dominant terms under the registered
  decomposition.  Restoring the native
  step envelope reduces the loss gap associated with step allocation, while the
  state-drift term records the changed future progress after the policies visit
  different states.  The latter offsets most of the recovered step cost.
  (c) Adaptive gain relative to the step-matched policy
  for every fixed depth, imbalance, and step-scale condition.  Colors denote
  depth, marker shapes denote the normalized-envelope numerator
  $s\in\{0.1,0.2\}$, and bars show 95\% trial-bootstrap intervals.  Positive
  values in this panel favor adaptive over step-matched and do not imply
  superiority to the native reciprocal learner.}
  \label{fig:cumulative}
\end{figure*}

The fixed design contains 20 trials, 12 depth--imbalance--step conditions, two
handedness replicates, and 30 epochs, giving 960 trajectories and 172800 updates.
Figure~\ref{fig:cumulative} asks whether locally accepted turns reduce the final
loss.  Panel (a) compares each nonreciprocal policy with the native reciprocal
learner; positive values mean that the nonreciprocal policy finishes with higher
loss.  The adaptive envelope narrows the mean final-loss gap from
$8.01\times10^{-3}$ to
$4.96\times10^{-3}$; the corresponding 95\% trial-cluster intervals are
$[5.59,10.63]\times10^{-3}$ and $[2.79,7.41]\times10^{-3}$.  Both gaps remain
positive in this experiment.  Panel (b) reports the dominant terms in the
registered decomposition.  Adaptive
stepping recovers $4.35\times10^{-2}$ of avoidable step-policy cost, but the
accompanying change in state drift offsets $92.9\%$ of that gain.  Panel (c)
confirms that the adaptive policy improves on the step-matched policy in every
tested condition; it does not compare either policy directly with the native
control.  The registered accounting identity closes within $1.28\times10^{-15}$.

\FloatBarrier
\section{Discussion}
\label{sec:discussion}

The main result is a hierarchy of constraints on physical learning.  Conservation
determines the feasible parameter motion.  Adjoint matching gives the physical
response a Gram form, while non-negative spectral feedback keeps the closed-loop
response within preconditioned gradient descent.  Antisymmetric boundary feedback
instead supplies an independent tangential direction while the selected-sample
error decreases instantaneously.
This distinction explains why local, normalized, and physically mediated updates
need not share the same optimization geometry.

This additional freedom can be beneficial only when it is coordinated with the
landscape.  In the tested linear transport family, the selected-sample curvature
ratio predicts the sign of the finite-step effect with high accuracy.  The cumulative experiment then exposes a
different bottleneck: accepting a turn changes the states at which all later
updates are evaluated.  Adaptive step control recovers most of the avoidable
step-policy cost, but state drift offsets most of that recovery.  The design
problem is therefore not to maximize rotation.  It is to introduce rotation where
the local geometry supports it and to control the stability of the trajectory
created afterward.

This hierarchy also determines fair baselines.  A reciprocal physical rule that
is algebraically equivalent to natural gradient or damped Gauss--Newton should be
compared with a matched metric-gradient control; its distinct physical value must
then be measured through implementation quantities such as solution time,
communication, energy, precision, robustness, or locality.  A nonreciprocal rule
requires the same comparison plus a sign mechanism for deciding when the
rotational component should be used.

The formal results apply to positive linear column-stochastic networks with fixed
support and matched adjoint response.  The three-port mixer is modeled as an
effective active boundary element, so a topology-local hardware realization
remains open.  Nonlinear conservative units, changing support, signed flows, and
noisy response solves provide concrete tests of how far the mechanism extends.

\section{Conclusion}

Conservative physical learning has two distinct geometric ingredients.
Conservation determines where parameters may move.  Within the matched response
class studied here, non-negative spectral feedback keeps the closed-loop dynamics
gradient-like, while an antisymmetric boundary component supplies a minimal
rotational degree of freedom.  Whether this freedom helps learning depends on
curvature over one step and on state drift over many steps.
By connecting exact response structure,
local sign prediction, and finite-horizon accounting, the framework turns
nonreciprocal learning into a testable design principle: introduce rotation where
the landscape supports it, and control how each turn changes the states visited
next.

\appendix

\section{Proofs of the response-kernel results}
\label{app:kernel}

For $p\in\Delta_d^\circ$ and a tangent vector $V$ with
$\one^{\mathsf T}V=0$, choose $s_i=V_i/p_i$.  Equation~\eqref{eq:q} then gives
\begin{equation}
 Q(p)s=V-p\sum_iV_i=V.
 \label{eq:app-replicator}
\end{equation}
Adding a constant to $s$ leaves this vector unchanged.  In reduced simplex
coordinates, the differences $s_i-s_d$ are the components of the covector dual to
$V$ under the Shahshahani metric.  The Poincar\'e lemma therefore gives a scalar
potential precisely when the one-form in Eq.~\eqref{eq:oneform} is closed.

For the layered network, a perturbation of column $p_i^{(\ell)}$ changes sample
$s$ by
\begin{equation}
 \delta F_s=x_{s,i}^{(\ell)}R^{(\ell)}\delta p_i^{(\ell)}.
 \label{eq:app-jacobian}
\end{equation}
Contracting two such Jacobian blocks with the local conductance
$\rho_\ell Q(p_i^{(\ell)})$ and summing over layers and columns gives
Eq.~\eqref{eq:blocks}.  Symmetry and positive semidefiniteness follow immediately
from $K_L=J_LM_LJ_L^{\mathsf T}$ and $M_L\succeq0$.

For spectral feedback, write $K=\sum_\lambda\lambda P_\lambda$ and use the same
eigenprojectors for $r(K)$.  This gives Eq.~\eqref{eq:dissipation}.  Moreover,
$H$ in Eq.~\eqref{eq:H} is symmetric positive semidefinite and
\begin{align}
 -H\nabla_\theta\cL
 &=MJ^{\mathsf T}\psi(K)JM J^{\mathsf T}e \nonumber\\
 &=MJ^{\mathsf T}\psi(K)K e
  =MJ^{\mathsf T}r(K)e=\dot\theta,
 \label{eq:app-preconditioner}
\end{align}
where the zero-eigenvalue contribution vanishes because
$MJ^{\mathsf T}P_{\ker K}=0$.

For the skew construction, $P$ is the identity and $C^2=-I$ on
$\one^\perp$.  Under $\operatorname{range}(K_\star)=\one^\perp$,
$K_\star K_\star^\dagger=P$ and $PT_{\alpha,\Omega}=T_{\alpha,\Omega}$.
This gives Eq.~\eqref{eq:spiral}.  A Riemannian
gradient linearization at a stationary point
has the form $-G^{-1}A$, with $G$ positive definite and $A$ symmetric.  It is
similar to the real symmetric matrix $-G^{-1/2}AG^{-1/2}$ and has a real
spectrum.  The pair $-\alpha\pm i\Omega$ therefore establishes the obstruction on
the locally identifiable quotient.

\section{Orientation-averaged finite-step expansion}
\label{app:finite}

For one probability column, let $\bar s=p^{\mathsf T}s$ and
$\operatorname{var}_p(s)=\sum_i p_i(s_i-\bar s)^2$.  The first two derivatives of
the exponential path in Eq.~\eqref{eq:retraction} are
\begin{align}
 A_p(s)_i&=p_i(s_i-\bar s),
 \label{eq:app-A}\\
 B_p(s)_i&=p_i\left[(s_i-\bar s)^2-
 \operatorname{var}_p(s)\right].
 \label{eq:app-B}
\end{align}
For $s^0+\sigma\Omega s^1$, the first derivative is
$U+\sigma\Omega V$.  Orientation averaging cancels terms odd in $\sigma$.
Relative to the $s^0$ path, the even second derivative of the prediction is
$\Omega^2H_V$, while the squared first prediction derivative gains
$\Omega^2\lVert G\rVert^2$.  Differentiating the mean squared loss twice and
applying Taylor's theorem yields
\begin{equation}
 \frac{\bar\cL_+(\eta)+\bar\cL_-(\eta)}{2}
 -\bar\cL_0(\eta)
 =\frac{\eta^2\Omega^2}{2N}
 \left(\lVert G\rVert^2-e^{\mathsf T}H_V\right)+O(\eta^3),
 \label{eq:app-Q}
\end{equation}
which is Eqs.~\eqref{eq:secondorder} and \eqref{eq:Q}.

\section{Definitions for the registered finite-horizon accounting identity}
\label{app:ledger}

Let $U_{t,a}(\theta;\eta)$ be the exponential update at time $t$ with score
$b_t(\theta)+a r_t(\theta)$, and define its dataset-loss increment by
\begin{equation}
 \delta_{t,a}(\theta;\eta)
 =\cL\bigl(U_{t,a}(\theta;\eta)\bigr)-\cL(\theta).
 \label{eq:app-delta}
\end{equation}
For $X\in\{\mathrm{NR},\mathrm R\}$, introduce the trajectory-specific shorthand
\begin{equation}
 \delta_{t,a}^{X}(\eta)
 :=\delta_{t,a}(\theta_t^{X};\eta),
 \qquad h_t^{X}:=h_t(\theta_t^{X}).
 \label{eq:app-shorthand}
\end{equation}
The coupled trajectories satisfy
$\theta_{t+1}^{\mathrm{NR}}
=U_{t,\sigma_t\gamma_t}(\theta_t^{\mathrm{NR}};\nu_t)$ and
$\theta_{t+1}^{\mathrm R}
=U_{t,0}(\theta_t^{\mathrm R};h_t^{\mathrm R})$, with
$\theta_0^{\mathrm{NR}}=\theta_0^{\mathrm R}$.  Here,
$\sigma_t\in\{-1,+1\}$, $\gamma_t\geq0$, and $\nu_t$ are the realized orientation,
rotation magnitude, and nonreciprocal step.  Define
\begin{align}
 \mathsf E_t={}&\frac12\left(
 \delta_{t,+\gamma_t}^{\mathrm{NR}}(\nu_t)
 +\delta_{t,-\gamma_t}^{\mathrm{NR}}(\nu_t)\right)
 -\delta_{t,0}^{\mathrm{NR}}(\nu_t),
 \label{eq:app-E}\\
 \mathsf H_t={}&\delta_{t,\sigma_t\gamma_t}^{\mathrm{NR}}(\nu_t)
 -\frac12\left(
 \delta_{t,+\gamma_t}^{\mathrm{NR}}(\nu_t)
 +\delta_{t,-\gamma_t}^{\mathrm{NR}}(\nu_t)\right),
 \label{eq:app-handedness}\\[0.5mm]
 \mathsf S_t={}&\delta_{t,0}^{\mathrm{NR}}(\nu_t)
 -\delta_{t,0}^{\mathrm{NR}}(h_t^{\mathrm{NR}}),
 \label{eq:app-step}\\[0.5mm]
 \mathsf D_t={}&\delta_{t,0}^{\mathrm{NR}}(h_t^{\mathrm{NR}})
 -\delta_{t,0}^{\mathrm R}(h_t^{\mathrm R}).
 \label{eq:app-drift}
\end{align}
The first three terms are evaluated at the nonreciprocal state, whereas
$\mathsf D_t$ compares native reciprocal progress from the two current states.
Adding the four differences gives
\begin{align}
 &\mathsf E_t+\mathsf H_t+\mathsf S_t+\mathsf D_t
 \nonumber\\[-0.3ex]
 &=\delta_{t,\sigma_t\gamma_t}^{\mathrm{NR}}(\nu_t)
 -\delta_{t,0}^{\mathrm R}(h_t^{\mathrm R}) \nonumber\\
 &=\bigl[\cL(\theta_{t+1}^{\mathrm{NR}})
 -\cL(\theta_t^{\mathrm{NR}})\bigr]
 -\bigl[\cL(\theta_{t+1}^{\mathrm R})
 -\cL(\theta_t^{\mathrm R})\bigr].
 \label{eq:app-telescope}
\end{align}
Because the trajectories share their initial state, summing over $t$ telescopes
to Eq.~\eqref{eq:ledger}.

\section*{Data availability}

The data tables and source code used to generate the results and figures are
contained in the reproducibility archive accompanying this manuscript.  A
public repository identifier will be added before submission.

\enlargethispage{3\baselineskip}
\bibliography{references}

\end{document}


\title{Supplemental Material for ``Reciprocity Separates Gradient Flow from Rotation in Conservative Physical Learning''}
\author{Ruiwu Niu}
\email{rniu@hksyu.edu}
\affiliation{Department of Data Science and Digital Innovation, Hong Kong Shue Yan University}

\author{Xiaowen Bi}
\email{xiaowenbi@bnbu.edu.cn}
\affiliation{Faculty of Science and Technology, Beijing Normal-Hong Kong Baptist University}

\author{Micha\"el Antonie van Wyk}
\affiliation{School of Electrical and Information Engineering, University of the Witwatersrand, Johannesburg, South Africa}

\date{August 30, 2026}
\maketitle

\section{Scope}

The main article contains the proofs needed to follow the central results.  This
supplement retains counterexample arithmetic, experimental algorithms,
registered denominators, and the reproducibility contract.

\section{Data-generating model}

Each task uses three unit-mass input vertices and three independently generated
simplex mixtures.  All trainable columns contain a fixed uniform component, so the
registered trajectories remain in the simplex interior.  A sample is selected by
the frozen task generator and supplies the boundary residual.  The reciprocal
score uses leakage-regularized response, while the nonreciprocal score adds the
unit skew direction.  Both scores are applied through the normalized exponential
map in the main article.

The software computes response quantities in two algebraically independent ways:
the explicit layer sum and a direct Jacobian--metric product.  The trajectory check
likewise compares a layer-sum update with an independently assembled
metric-Gauss--Newton update.  These comparisons identify implementation errors;
their numerical residuals are distinct from trial-to-trial statistical
uncertainty.

\section{Frozen structural reproduction}

The structural design was frozen with root seed 20261020.  It retains task-family
identifiers 0--24 at depths $2$, $3$, $4$, and $6$.  Every hidden and output width
is three.  The rotation ratios are
$\Omega/\alpha\in\{0,0.5,1,2\}$ with $\alpha=1$.  Reciprocal trajectories contain
eight updates with step $0.05$ and leakage $0.1$.  The registered denominator is
100 family--depth pairs, 400 mode rows, 900 trajectory rows, four disconnected
controls, and 14 explicit-curvature rows.

The block was required to satisfy residual thresholds of $10^{-10}$ for response
identities, $10^{-12}$ for symmetry and column conservation, rank six for every
connected joint operator, rank two for every selected-sample operator, and rank
one for each disconnected control.  The two exact curvature examples were tested
at
\begin{equation}
 \eta\in\{10^{-2},5\!\times\!10^{-3},2\!\times\!10^{-3},10^{-3},
 5\!\times\!10^{-4},2\!\times\!10^{-4},10^{-4}\}.
\end{equation}
Their raw even loss differences were required to have fitted log--log slopes in
$[1.9,2.1]$ and relative coefficient error below $10^{-4}$ at the smallest step.

The two exact one-layer examples use a unit scalar input and
$\alpha=\Omega=1$.  The negative example starts from
$p=(0.1,0.1,0.8)^{\mathsf T}$ with target
$y=(0.3,0.4,0.3)^{\mathsf T}$ and gives
$\cQ=-0.9271875$.  The positive example starts from
$p=(1/3,1/3,1/3)^{\mathsf T}$ with target
$y=(0.5,0.3,0.2)^{\mathsf T}$ and gives
$\cQ=0.026666\ldots$.

\begin{table}[h]
\caption{Frozen structural-block outcome.  The whole block retains status FAIL
because the rank gate is conjunctive.}
\begin{ruledtabular}
\begin{tabular}{lc}
Quantity & Outcome \\
\hline
Maximum factorization residual & $1.208\times10^{-16}$ \\
Maximum score-response residual & $1.004\times10^{-15}$ \\
Maximum trajectory discrepancy & $2.220\times10^{-16}$ \\
Maximum complex-pair residual & $3.935\times10^{-15}$ \\
Maximum dissipation residual & $1.636\times10^{-15}$ \\
Minimum finite weight & $8.397\times10^{-2}$ \\
Rank six, depths 2/3/4 & $25/25$ at each depth \\
Rank six, depth 6 & $8/25$ \\
Negative-$\cQ$ slope & $1.999921$ \\
Positive-$\cQ$ slope & $1.999405$ \\
Registered block status & \textbf{FAIL: joint rank} \\
\end{tabular}
\end{ruledtabular}
\end{table}

The 17 rank-five depth-6 operators have retained-spectrum condition numbers from
approximately $3.80\times10^6$ to $5.53\times10^9$.  All factorization rows remain
at floating-point closure.  The outcome is therefore a numerical identifiability
failure under the frozen rank tolerance, while preserving the exact Gram identity.
The disconnected controls retain rank one and return zero response along the
registered intercomponent null mode.

\section{Disjoint local-threshold validation}

The predictor and validation population were frozen before execution.  The data
root seed is 20261021 and the bootstrap seed is 20261022.  Fifty new task families
are evaluated on architectures L1, L2-b3, L4-b3, and L6-b3 at imbalance strengths
$\beta\in\{0,1,2,4,8\}$.  The 1000 configuration rows are correlated within task;
all reported intervals use 5000 task-family bootstrap resamples.

The frozen rule predicts $\cQ<0$ exactly when the selected-sample local curvature
ratio exceeds one.  Acceptance requires cluster-bootstrap lower bounds above 0.90
for AUC and 0.85 for accuracy, a beta-zero favorable-fraction upper bound below
0.10, a beta-eight lower bound above 0.75, analytic/finite sign agreement of at
least 0.99, and the registered response and conservation tolerances.

\begin{table}[h]
\caption{Disjoint local-threshold validation.}
\begin{ruledtabular}
\begin{tabular}{lcc}
Statistic & Estimate & 95\% task-cluster interval \\
\hline
AUC & $0.997742$ & $[0.996364,0.998900]$ \\
Accuracy & $0.971$ & $[0.962,0.980]$ \\
$\Pr(\cQ<0\mid\beta=0)$ & $0$ & $[0,0]$ \\
$\Pr(\cQ<0\mid\beta=8)$ & $0.945$ & $[0.910,0.975]$ \\
Analytic/finite sign agreement & $1.000$ & N/A \\
Registered block status & \textbf{PASS} & N/A \\
\end{tabular}
\end{ruledtabular}
\end{table}

The maximum response residual is $1.072\times10^{-14}$, the maximum relative
coefficient residual is $2.188\times10^{-6}$, the maximum column residual is
$2.220\times10^{-16}$, and the minimum weight is $6.649\times10^{-3}$.  The
operator condition-number median is 2.960, with interquartile range
$[1.461,5.187]$ and maximum 89.015.

\section{Registered finite-horizon accounting experiment}

The finite-horizon experiment uses data seed 20260815, bootstrap seed 20260816,
20 trials, 30 epochs, two handedness replicates, and 12 combinations of depth,
imbalance, and step scale.  Handedness replicates are averaged within trial and
condition before resampling the 20 trials.  The registered tables contain 960
trajectories and 172800 primitive update rows.

The registered accounting identity is evaluated step by step before aggregation.  Maximum step,
trajectory, and paired-policy closure residuals are $1.292\times10^{-16}$,
$1.193\times10^{-15}$, and $6.423\times10^{-16}$, respectively.  The native
control is identical across paired policies and signs.  The minimum finite weight
is $6.889\times10^{-4}$.

\begin{table}[h]
\caption{Cumulative loss gaps and dominant accounting terms.  Positive final gaps
favor the native reciprocal control.}
\begin{ruledtabular}
\begin{tabular}{lrr}
Quantity & Step-matched & Adaptive envelope \\
\hline
Mean final gap to native & $8.00995\times10^{-3}$ & $4.96231\times10^{-3}$ \\
95\% trial-cluster interval & $[5.59,10.63]\times10^{-3}$ & $[2.79,7.41]\times10^{-3}$ \\
Step-policy contribution & $1.09871\times10^{-1}$ & $6.64061\times10^{-2}$ \\
State-drift contribution & $-1.01247\times10^{-1}$ & $-6.08903\times10^{-2}$ \\
\end{tabular}
\end{ruledtabular}
\end{table}

The adaptive envelope improves on the step-matched policy in all 12 conditions.
The mean gain is $3.04763\times10^{-3}$ with bootstrap interval
$[2.23560,3.94591]\times10^{-3}$.  Both policies retain a positive final gap to
the native reciprocal trajectory.  Under the registered decomposition, the
change in state drift offsets 92.85\% of the recovered step-policy gain.  All
figures are regenerated from immutable
tables.  Both new blocks have frozen designs, source-hash locks, complete
denominators, manifests, and independent verifiers; the registered structural FAIL
and local-threshold PASS reproduce exactly, and failed rows remain in place.